\documentclass{article}
\usepackage{nips15submit_e,times}
\usepackage[colorlinks,
           linkcolor=red,
           citecolor=blue,
           urlcolor=magenta,
           linktocpage,
           plainpages=false]{hyperref}
\usepackage{url}
\usepackage{algorithm}
\usepackage{algorithmic}
\usepackage{amsthm}
\usepackage{amsmath}
\usepackage{amssymb}
\usepackage{bm}
 \usepackage[numbers]{natbib}
\usepackage{wrapfig}
\DeclareMathOperator*{\argmax}{arg\,max}
\DeclareMathOperator*{\argmin}{arg\,min}

\newtheorem{assumption}{Assumption}

\newtheorem{theorem}{Theorem}
\newtheorem{definition}{Definition}

\newcommand{\PP}{\mathbb{P}}

\newcommand{\EE}[1]{\mathbb{E}\left[#1\right]}

\newcommand{\EEc}[2]{\mathbb{E}\left[#1\left|#2\right.\right]}

\newcommand{\CommaBin}{\mathbin{\raisebox{0.5ex}{,}}}
\newcommand*{\eqdef}{\triangleq}

\newcommand{\cO}{\mathcal{O}}

\renewcommand{\epsilon}{\varepsilon}

\newcommand{\nothere}[1]{}

\usepackage[disable]{todonotes}
\definecolor{graphicbackground}{rgb}{0.96,0.96,0.8}
\definecolor{rouge1}{RGB}{226,0,38}  
\definecolor{orange1}{RGB}{243,154,38}  
\definecolor{jaune}{RGB}{254,205,27}  
\definecolor{blanc}{RGB}{255,255,255} 
\definecolor{rouge2}{RGB}{230,68,57}  
\definecolor{orange2}{RGB}{236,117,40}  
\definecolor{taupe}{RGB}{134,113,127} 
\definecolor{gris}{RGB}{91,94,111} 
\definecolor{bleu1}{RGB}{38,109,131} 
\definecolor{bleu2}{RGB}{28,50,114} 
\definecolor{vert1}{RGB}{133,146,66} 
\definecolor{vert3}{RGB}{20,200,66} 
\definecolor{vert2}{RGB}{157,193,7} 
\definecolor{darkyellow}{RGB}{233,165,0}  
\definecolor{lightgray}{rgb}{0.9,0.9,0.9}
\definecolor{darkgray}{rgb}{0.6,0.6,0.6}
\definecolor{babyblue}{rgb}{0.54, 0.81, 0.94}
\definecolor{citrine}{rgb}{0.89, 0.82, 0.04}
\definecolor{misogreen}{rgb}{0.25,0.6,0.0}
\definecolor{PalePurp}{rgb}{0.66,0.57,0.66}
\definecolor{todocolor}{rgb}{0.66,0.99,0.99}
\definecolor{pearOne}{HTML}{2C3E50}
\definecolor{pearTwo}{HTML}{A9CF54}
\definecolor{pearTwoT}{HTML}{C2895B}
\definecolor{pearThree}{HTML}{E74C3C}
\colorlet{titleTh}{pearOne}
\colorlet{bull}{pearTwo}
\definecolor{pearcomp}{HTML}{B97E29}
\definecolor{pearFour}{HTML}{588F27}
\definecolor{pearFith}{HTML}{ECF0F1}
\definecolor{pearDark}{HTML}{2980B9}
\definecolor{pearDarker}{HTML}{1D2DEC}

\hypersetup{
	colorlinks,
	citecolor=pearDark,
	linkcolor=pearThree,
	urlcolor=pearDarker}

\newcommand{\SHOO}{\texttt{\textcolor[rgb]{0.5,0.2,0}{POO}}}
\newcommand{\StoSOO}{\texttt{StoSOO}}
\newcommand{\SOO}{\texttt{SOO}}
\newcommand{\HOO}{\texttt{HOO}}
\newcommand{\HCT}{\texttt{HCT}}
\newcommand{\UCT}{\texttt{UCT}}
\newcommand{\Zooming}{\texttt{TaxonomyZoom}}
\newcommand{\ATB}{\texttt{ATB}}
\title{Black-box optimization of noisy functions with unknown smoothness}

\author{Jean-Bastien Grill   \hspace{4em}  Michal Valko\\ SequeL team, INRIA Lille - Nord Europe, France\\ \texttt{\small jean-bastien.grill@inria.fr  \hspace{.5em}michal.valko@inria.fr}
\And \hspace{-3em}  R\'emi  Munos\\ \hspace{-3em} Google DeepMind, UK\thanks{on leave from SequeL team, INRIA Lille - Nord Europe, France}\\ \hspace{-3em}\texttt{\small munos@google.com}}
 \nipsfinalcopy 
\begin{document}
\maketitle
We study the problem of black-box optimization of a function $f$ of any dimension, given function evaluations perturbed by noise. The function is assumed to be locally smooth around one of its global optima, but this \emph{smoothness is unknown}. Our contribution is an adaptive optimization algorithm, \SHOO{} or
\textit{parallel optimistic optimization}, that is able to deal with this setting. \SHOO{} performs almost as well as the best known algorithms requiring the knowledge of the smoothness. Furthermore, \SHOO{} works for a larger class of functions than what was previously considered, especially for functions that are \emph{difficult to optimize}, in a very precise sense. We provide a \emph{finite-time} analysis of  \SHOO{}'s performance, which shows that its error after $n$ evaluations is at most a factor of $\sqrt{\ln n}$ away from the error of the best known optimization algorithms using the knowledge of the smoothness. 

\section{Problem definition}\label{sec:intro}
We treat the problem of optimizing a function $f: \mathcal{X} \rightarrow \mathbb{R}$ given a finite budget of $n$ noisy evaluations. We consider that the cost of any of these \textit{function evaluations} is high. That means, we care about assessing the optimization performance in terms of the sample complexity, i.e.,\@ the number of~$n$ function evaluations. This is typically the case when one needs to tune parameters for a complex system seen as a black-box, which performance can only be evaluated by a costly simulation. One such example is the \textit{hyper-parameter tuning} where the sensitivity to perturbations
is large and the derivatives of the objective function with respect to these parameters
do not exist or are unknown.

Such setting fits the sequential decision-making setting under \textit{bandit feedback}.
In this setting, the actions are the points that lie in a domain~$\mathcal{X}$.
At each step $t$, an algorithm selects an action $x_t\in\mathcal{X}$ and receives a reward $r_t$, 
which is a noisy function evaluation such that $r_t =  f(x_t) + \varepsilon_t$, 
where $\varepsilon_t$ is a bounded noise with $\EEc{\varepsilon_t}{x_t} = 0$. After $n$ evaluations,
the algorithm outputs its best guess $x(n)$, which can 
be different from $x_n$. The performance measure we want to minimize is the value of the function at the returned point compared to the optimum, also referred to as \textit{simple regret},
\[
R_n \eqdef \sup_{x\in \mathcal{X}}f(x)  - f\left(x\left(n\right)\right)\!.
\]
We assume there exists at least one point $x^\star\in \mathcal{X}$ such that $f(x^\star)=\sup_{x\in \mathcal{X}}f(x)$.

The relationship with bandit settings motivated \UCT{}~\cite{kocsis2006bandit,coquelin2007bandit}, an empirically successful heuristic that hierarchically partitions domain~$\mathcal{X}$ and selects the next point $x_t\in\mathcal{X}$ using upper confidence bounds~\cite{auer2002finite}.
The empirical success of \UCT{} on one side but the absence of performance guarantees for it on the other, incited research 
on similar but theoretically founded algorithms~\cite{bubeck2011x,kleinberg2008multi,munos2014from,azar2014online,bull2015adaptive}.

As the global optimization of the unknown function without absolutely any assumptions would be 
a daunting needle-in-a-haystack problem, most of the algorithms assume at least a very weak assumption 
that the function does \textit{not decrease faster than a known rate} around \textit{one} of its global optima.
In other words, they assume a certain \textit{local smoothness} property of $f$.
This smoothness is often expressed in the form of a semi-metric $\ell$
that quantifies this regularity~\cite{bubeck2011x}.
Naturally, this regularity also influences the guarantees that these
algorithms are able to furnish. Many of them define 
a \textit{near-optimality dimension} $d$ or a \textit{zooming dimension}.
These are $\ell$-dependent quantities used to bound the simple regret $R_n$ or a related notion called \emph{cumulative regret}.

Our work focuses on a notion of such near-optimality dimension $d$ that does not directly relate the smoothness property of $f$ to a specific metric $\ell$ but \emph{directly} to the \emph{hierarchical partitioning} $\mathcal{P}=\{\mathcal{P}_{h,i}\}$, a\emph{ tree-based representation} of the space used by the algorithm.
Indeed, an interesting fundamental question is to determine a good characterization of the difficulty of the optimization for an algorithm that uses a given hierarchical partitioning of the space $\mathcal{X}$ as its input. 
The kind of hierarchical partitioning $\{\mathcal{P}_{h,i}\}$ we consider is similar to the ones introduced in prior work: for any depth $h\geq 0$ in the tree representation, the set of \emph{cells} $\{\mathcal{P}_{h,i}\}_{1\leq i\leq I_h}$ form a partition of $\mathcal{X}$, where $I_h$ is the number of cells at depth $h$. At depth 0, the root of the tree, there is a single cell $\mathcal{P}_{0,1}=\mathcal{X}$. A cell $\mathcal{P}_{h,i}$ of depth $h$ is split into several children subcells $\{\mathcal{P}_{h+1,j}\}_{j}$ of depth $h+1$. We refer to the standard partitioning as to one where each cell is split into regular same-sized subcells~\cite{preux2014bandits}.

An important insight, detailed in Section~\ref{sec:bck},
is that a near-optimality dimension $d$ that is independent from the partitioning used by an algorithm 
(as defined in prior work~\cite{bubeck2011x,kleinberg2008multi,azar2014online}) \textit{does not embody the optimization difficulty perfectly}.\/ 
This is easy to see, as for any $f$ we could 
define a  partitioning, perfectly suited for $f$. 
An example is a partitioning, that at the root splits 
$\mathcal{X}$ into $\{x^\star\}$ and $\mathcal{X}\setminus x^\star$,
which makes the optimization trivial, whatever $d$ is.
This insight was already observed by~\citet{slivkins2011multi-armed} and~\citet{bull2015adaptive},
whose \textit{zooming dimension} depends both 
on the function and the partitioning. 

In this paper, we define a notion of near-optimality dimension $d$ which measures the complexity of the optimization problem \emph{directly in terms of the partitioning} used by an algorithm. First, we make the following local smoothness assumption about the function, expressed in terms of the partitioning  and \emph{not any metric}: For a given partitioning $\mathcal{P}$, we assume that there exist $\nu>0$ and $\rho\in(0,1)$, s.t.\@,
\begin{align*}\label{ass1}
\forall h\geq 0, \forall x\in\mathcal{P}_{h,i^\star_h},\quad  f(x)\ge f\left(x^\star\right)-\nu\rho^h,
\end{align*}
where $\left(h,i^\star_h\right)$ is the (unique) cell of depth $h$ containing $x^\star$. 
Then, we define the near-optimality dimension $d(\nu,\rho)$ as
\[d(\nu, \rho) \eqdef \inf \left\{d'\in\mathbb{R}^+:\exists C>0, \forall h\geq 0, \mathcal{N}_h(2\nu\rho^h) \le C\rho^{-d'h}\right\},\!\]
where for all $\varepsilon > 0$, $\mathcal{N}_h(\varepsilon)$ is the number of cells $\mathcal{P}_{h,i}$ of depth $h$ s.t. $\sup_{x\in\mathcal{P}_{h,i}} f(x)\ge f\left(x^\star\right)-\varepsilon$. 
Intuitively, functions with smaller $d$ are easier to optimize and we denote $(\nu,\rho)$, for which  $d(\nu,\rho)$ is the smallest, 
as $(\nu_\star,\rho_\star)$. Obviously, $d(\nu,\rho)$ depends on $\mathcal{P}$ and $f$,  but \emph{does not depend} on any choice of a specific metric. In Section~\ref{sec:bck}, we argue that this definition of~$d$\footnote{we use the simplified notation $d$ instead of $d(\nu,\rho)$ for clarity when no confusion is possible} encompasses the optimization complexity \emph{better}. We stress this is not an artifact of our analysis and previous algorithms, such as \HOO{}~\cite{bubeck2011x},  \Zooming{}~\cite{slivkins2011multi-armed}, or \HCT{}~\cite{azar2014online}, can be shown to scale with this new notion of~$d$.
%

Most of the prior bandit-based algorithms proposed for function optimization, for either deterministic or stochastic setting, assume that the smoothness of the 
optimized function is \textit{known}. This is the 
case of known \emph{semi-metric}~\cite{bubeck2011x,azar2014online} and
\emph{pseudo-metric}~\cite{kleinberg2008multi}.
This assumption limits the application of these algorithms and opened a very compelling question of whether this knowledge is necessary. 

Prior work responded with algorithms not requiring this knowledge. \citet{bubeck2011lipschitz} provided an algorithm for optimization
of Lipschitz functions without the knowledge of the Lipschitz constant. However, they have to assume that $f$ is twice differentiable and a bound on the second order derivative is known. 
\citet{combes2014unimodal} treat unimodal $f$ restricted to dimension one. 
\citet{slivkins2011multi-armed} considered a general optimization problem embedded in a \textit{taxonomy}\footnote{which is similar to the hierarchical partitioning previously defined} and provided guarantees as a function of the \textit{quality} of the taxonomy. The quality refers to the probability of reaching two cells belonging to the same branch that can have values that differ by more than half of the diameter (expressed by the true metric) of the branch. The problem is that the algorithm needs a lower bound on this quality (which can be tiny) and the performance depends inversely on this quantity. Also it assumes that the quality is strictly positive. In this paper, we do not rely on the knowledge of quality and also consider a more general class of functions for which the quality can be~$0$ (Appendix~\ref{ss:quality}).

Another   direction has been followed by \citet{munos2011optimistic}, where in the deterministic case (the function evaluations are not perturbed by noise), their \SOO{} algorithm performs almost as well as the best known algorithms without the knowledge of the function smoothness. \SOO{} was later extended to \StoSOO{}~\cite{valko2013stochastic} for the stochastic case. However \StoSOO{} only extends \SOO{} for a limited case of \textit{easy instances} of functions for which there exists a semi-metric under which $d=0$.
Also, \citet{bull2015adaptive} provided a similar simple regret bound for \ATB{}  for a class of functions, called \textit{zooming continuous functions}, which is related to the class of functions for which there exists a semi-metric under which the near-optimality dimension is $d=0$. But none of the prior work considers a more general class of functions 
where there is no semi-metric adapted to the standard partitioning  for which $d=0$. 

\begin{figure}
\begin{center}
\vspace{-1.5em}
\includegraphics[width=0.45\textwidth]{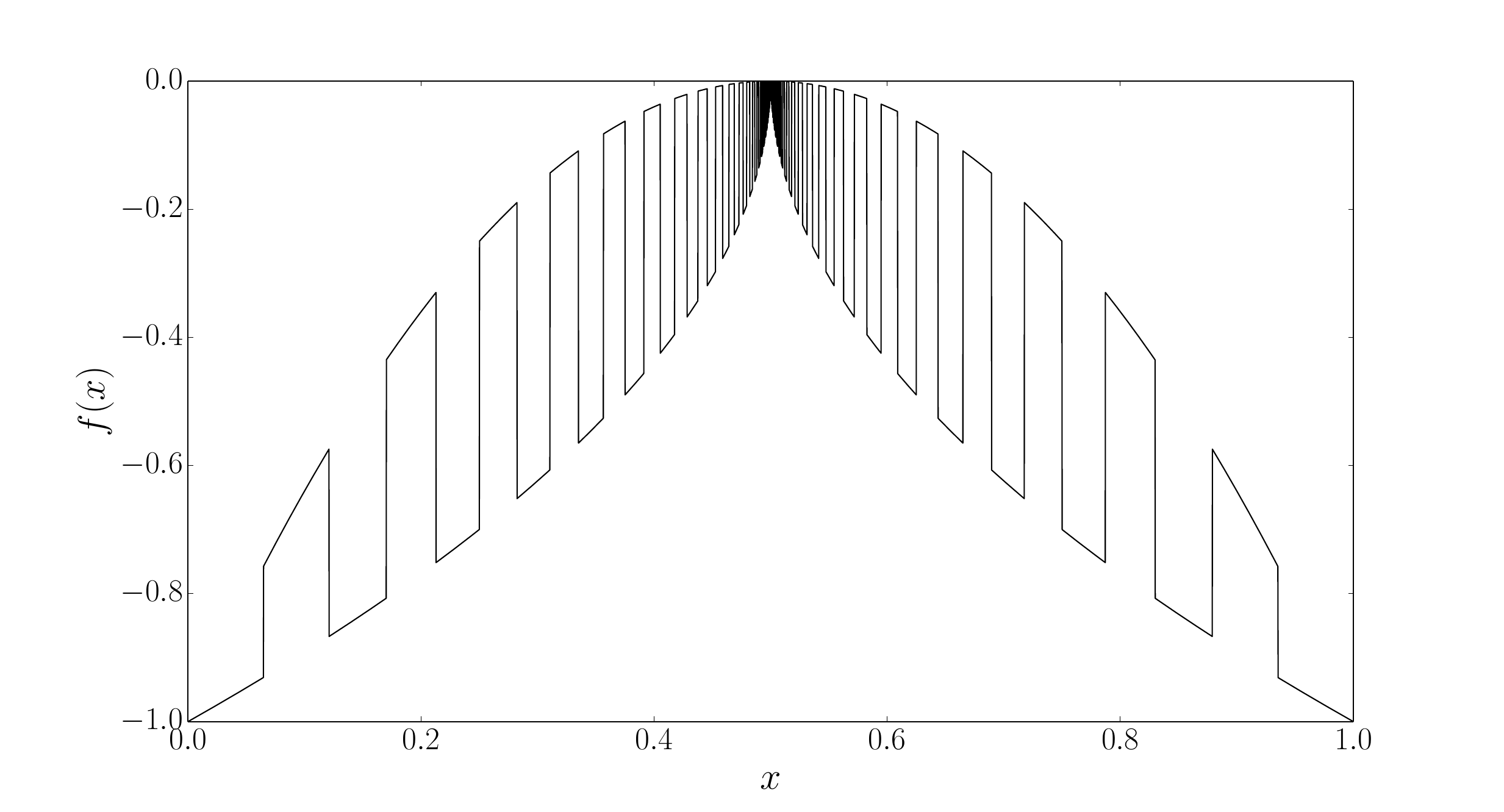}
\includegraphics[width=0.45\textwidth]{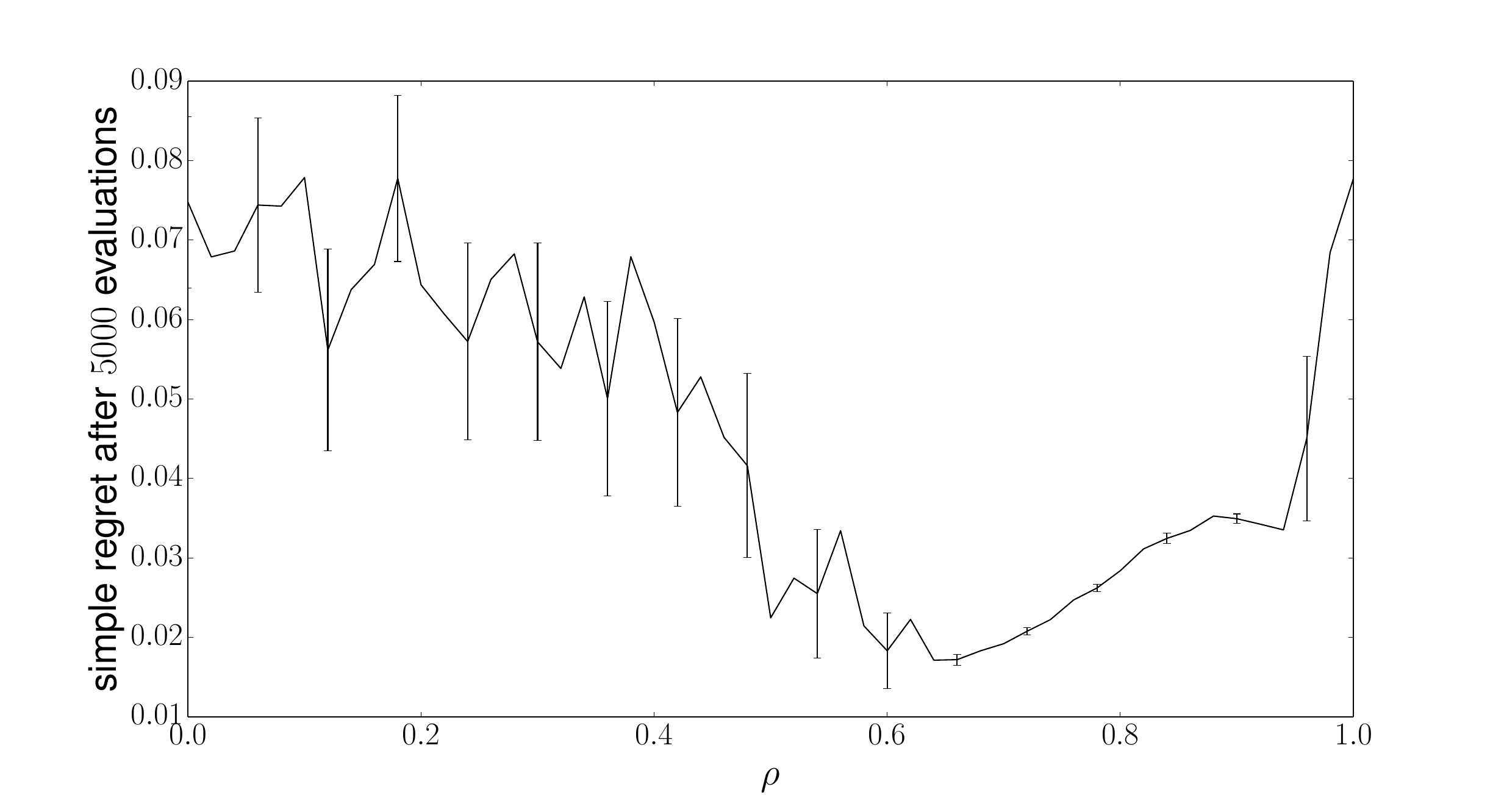}
\vspace{-1em}
\caption{ 
Difficult function
$f:x\rightarrow s\left(\log_2|x-0.5|\right)\cdot(\sqrt{|x-0.5|} - {(x-0.5)}^2)-\sqrt{|x-0.5|}$ where, $s(x) = 1$ if the fractional part of $x$, that is,  $x - \lfloor x \rfloor$,  is in $[0,0.5]$ and $s(x) = 0$,  if it is in $(0.5,1)$.
\emph{Left:} Oscillation between two envelopes of different smoothness leading to a nonzero $d$ for a standard partitioning. 
\emph{Right:} Simple regret of \HOO{} after $5000$ evaluations for different values of $\rho$. }
\vspace{-1.5em}
\end{center}
\label{fig:regretrho}
\end{figure}

To give an example of a difficult function, consider the function in Figure~\ref{fig:regretrho}.  
It possesses a lower and upper envelope around its global optimum that are equivalent to $x^2$ and $\sqrt{x}$; and therefore have different smoothness. 
Thus, for a standard partitioning, there is no 
semi-metric of the form $\ell(x,y) = ||x-y||^\alpha$ 
for which the near-optimality dimension is $d=0$, as shown by~\citet{valko2013stochastic}.
Other examples of nonzero near-optimality dimension are the functions that 
for a standard partitioning behave differently depending on the direction, for instance $f:(x,y)\mapsto 1 - |x| - y^2$. 

Using a bad value for the $\rho$ parameter can have dramatic consequences on the simple regret. In Figure~\ref{fig:regretrho}, we show the simple regret after $5000$ function evaluations for different values of $\rho$. For the values of $\rho$ that are too low, the algorithm does not explore enough and is stuck in a local maximum while for values of $\rho$ too high the algorithm wastes evaluations by exploring too much. 

%
In this paper, we provide a new algorithm, \SHOO{}, \textit{parallel optimistic optimization}, which competes with the best algorithms that assume the knowledge of the function smoothness, for a larger class of functions than was previously done. Indeed, \SHOO{} handles a panoply of functions, including \textit{hard instances}, i.e., such that $d>0$, like the function illustrated above. We also recover the result of \StoSOO{} and \ATB{} for functions with $d=0$. 
In particular, we bound the \SHOO{}'s simple regret as
\[\mathbb{E}[R_n] \leq \cO\left(\left(\left(\ln^2n\right)/n\right)^{1/(2+d(\nu_\star,\rho_\star))}\right)\!.\]
This result should be compared to the simple regret of the best known algorithm that uses the knowledge of the metric under which the function is smooth, or equivalently $(\nu,\rho)$, which is of the order of $\cO((\ln n/n)^{1/(2+d)})$.
Thus \SHOO{}'s performance is at most a factor of $(\ln n)^{1/(2+d)}$ away from that of the best known optimization algorithms that require the knowledge of the function smoothness. Interestingly, this factor decreases with the complexity measure $d$: the harder the function to optimize, the less important it is to know its precise smoothness.


%



%


\section{Background and assumptions}\label{sec:bck}
\subsection{Hierarchical optimistic optimization}
\label{ss:hoo}

\SHOO{}{} optimizes functions \textit{without} the knowledge of their smoothness using a subroutine, an anytime algorithm optimizing functions \textit{using} the knowledge of their smoothness.  
In this paper, we use a modified version of \HOO{}~\cite{bubeck2011x} as such subroutine. Therefore, we embark with a quick review of \HOO{}.

\HOO{} follows an optimistic strategy close to \UCT{}~\cite{kocsis2006bandit}, but unlike \UCT{}, it uses proper confidence bounds to provide theoretical guarantees. \HOO{} refines a partition of the space based on a hierarchical partitioning, where at each step, a yet unexplored cell (a leaf of the corresponding tree) is selected, and the function is evaluated at a point within this cell.
The selected path (from the root to the leaf) is the one that maximizes the minimum value $U_{h,i}(t)$ among all cells of each depth, where the value $U_{h,i}(t)$ of any cell $\mathcal{P}_{h,i}$ is defined as
\[U_{h,i}(t) = \widehat{\mu}_{h,i}(t) + \sqrt{\frac{2\ln(t)}{N_{h,i}(t)}}+\nu\rho^h,\]
where $t$ is the number of evaluations done so far, $\widehat{\mu}_{h,i}(t)$ is the empirical average of all evaluations done within  $\mathcal{P}_{h,i}$, and $N_{h,i}(t)$ is the number of them. 
The second term in the definition of $U_{h,i}(t)$ is a Chernoff-Hoeffding type confidence interval, measuring the estimation error induced by the noise. The third term, $\nu \rho^h$ with $\rho\in(0,1)$ is, by assumption, a bound on the difference $f(x^\star)-f(x)$ for any $x\in\mathcal{P}_{h,i^\star_{h}}$, a cell containing $x^\star$. It is this bound, where \HOO{} relies on the knowledge of the smoothness, because the algorithm requires the values of $\nu$ and $\rho$. In the next sections, we clarify the assumptions made by \HOO{} vs.\@\,related algorithms and point out the differences with \SHOO{}.

\subsection{Assumptions made in prior work}
\label{ss:prior}
Most of previous work relies on the knowledge of a semi-metric on $\mathcal{X}$ such that the function is 
either locally smooth near one of its maxima with respect to this metric 
~\cite{munos2011optimistic,valko2013stochastic,azar2014online}
or require a stronger, weakly-Lipschitz assumption~\cite{bubeck2011x,munos2014from,azar2014online}.
Furthermore, \citet{kleinberg2008multi} assume the full metric. 
Note, that the semi-metric does not require the triangular inequality to hold. For instance, consider the semi-metric $\ell(x,y) = ||x-y||^\alpha$ on $\mathbb{R}^p$ with $||\cdot||$ being the Euclidean metric. When $\alpha < 1$ then this semi-metric does not satisfy the triangular inequality. However, it is a metric for $\alpha\ge1$. Therefore, using only semi-metric  allows us to consider a larger class of functions. 

Prior work typically requires two assumptions. The first one is on semi-metric $\ell$
and the function.
An example is the \textit{weakly-Lipschitz} assumption needed by~\citet{bubeck2011x} which requires that
\[\forall x,y\in\mathcal{X},\quad  f(x^\star) - f(y) \le f(x^\star) - f(x) + \max \left\{f(x^\star)  - f(x), \ell\left(x,y\right)\right\}\!.\]
It is a weak version of a Lipschitz condition, restricting $f$ in particular for the values close to~$f(x^\star)$. 

More recent results~\cite{munos2011optimistic,valko2013stochastic,azar2014online} assume only a \textit{local smoothness}  around one of the function maxima,
\[x\in\mathcal{X}\quad f(x^\star) - f(x) \le \ell(x^\star,x).\]

The second common assumption \textit{links} the hierarchical partitioning  with the semi-metric. It requires the partitioning to be \textit{adapted} to the (semi) metric.  More precisely the well-shaped assumption states that there exist $\rho < 1$ and $\nu_1 \ge \nu_2 > 0$, such that for any depth $h \ge 0$ and index $i = 1,\dots,I_h$, the subset $\mathcal{P}_{h,i}$ is \emph{contained by} and \emph{contains} two open balls of radius $\nu_1\rho^h$ and $\nu_2\rho^h$ respectively, where the balls are w.r.t.~the same semi-metric used in the definition of the function smoothness.

`Local smoothness' is weaker than `weakly Lipschitz' and therefore preferable. 
Algorithms requiring the local-smoothness assumption always sample a cell $\mathcal{P}_{h,i}$ in a special \textit{representative point} and, in the stochastic case, collect several function evaluations from the same point before splitting the cell. 
This is not the case of \HOO{}, which allows  to sample {\em any} point inside the selected cell and to expand each cell after one sample. This additional flexibility comes at the price of requiring the stronger weakly-Lipschitzness assumption. 
Nevertheless, although \HOO{} does not wait before expanding a cell, it does something similar by selecting a path from the root to this leaf that maximizes the minimum of the $U$-value over the cells of the path, as mentioned in Section~\ref{ss:hoo}. The fact that \HOO{} follows an optimistic strategy even after reaching the cell that possesses the minimal $U$-value along the path is not used in the analysis of the \HOO{} algorithm. 


Furthermore, a reason for better dependency on the smoothness in other algorithms, e.g.,  \HCT{}~\cite{azar2014online}, is not only algorithmic:  \HCT{} needs to assume a slightly stronger condition on the cell, i.e., that the single center of the two balls (one that covers and the other one that contains the cell) is actually the same point that \HCT{} uses for sampling. This is stronger than just assuming that there simply exist such centers of the two balls, which are not necessarily the same points where we sample (which is the \HOO{} assumption).  Therefore, this is \emph{in contrast with \HOO{}} that samples \emph{any point} from the cell. 
In fact, it is straightforward to modify \HOO{} to only sample at a representative point in each cell
and only require the local-smoothness assumption. In our analysis and the algorithm, we 
use this modified version of \HOO{}, thereby profiting from this weaker assumption.


Prior work~\cite{kleinberg2008multi,bubeck2011x,munos2011optimistic,azar2014online,munos2014from} often defined some `dimension' $d$ of the near-optimal space of $f$ measured according to the (semi-) metric $\ell$. For example, the so-called \textit{near-optimality dimension} \cite{bubeck2011x} measures the size of the near-optimal space $\mathcal{X}_\varepsilon = \{ x \in \mathcal{X}: f(x) > f(x^\star) - \varepsilon\}$ in terms of \textit{packing numbers}:
For any $c>0, \varepsilon_0 >0$, the $(c, \varepsilon_0)$-near-optimality dimension $d$
of~$f$ with respect to $\ell$ is defined as
\begin{equation}\label{eq:near-opt}
\inf \left\{d\in[0,\infty): \exists C \text{ s.t. }
\forall\varepsilon\le\varepsilon_0\text{, }\mathcal{N}(\mathcal{X}_{c\varepsilon},\ell,\varepsilon)\le
C\varepsilon^{-d}\right\}\!,
\end{equation}
where for any subset $A\subseteq\mathcal{X}$, the packing number $\mathcal{N}(A,\ell,\varepsilon)$ is the maximum number of disjoint balls of radius $\varepsilon$ contained in $A$.

\subsection{Our assumption}
\label{ss:ourass}

Contrary to the previous approaches, we need \emph{only a single assumption}. We do not introduce any (semi)-metric and instead directly relate $f$ to the hierarchical partitioning $\mathcal{P}$, defined in Section~\ref{sec:intro}.
Let $K$ be the maximum number of children cells $(\mathcal{P}_{h+1,j_k})_{1\leq k\leq K}$ per cell $\mathcal{P}_{h,i}$. 
We remind the reader that given a global maximum $x^\star$ of $f$,  $i^\star_h$ denotes the index of the unique cell of depth $h$ containing $x^\star$, i.e.,\@ such that $x^\star\in \mathcal{P}_{h,i^\star_h}$.  
With this notation we can  state our sole assumption on both the  partitioning $(\mathcal{P}_{h,i})$ and the function~$f$. 
\begin{assumption}
\label{ass:lashoo}
There exists $\nu>0$ and $\rho\in(0,1)$ such that
\[\forall h\geq 0, \forall x\in\mathcal{P}_{h,i^\star_h}, \quad f(x)\ge f\left(x^\star\right)-\nu\rho^h.\]
\end{assumption}

The values $(\nu,\rho)$ defines a lower bound on the possible drop of $f$ near the optimum $x^\star$ according to the partitioning. The choice of the exponential rate $\nu\rho^h$ is made to cover a very large class of functions, as well as to relate to results from prior work. In particular, for a standard partitioning on $\mathbb{R}^p$ and any $\alpha,\beta>0$, any function $f$ such that $f(x)\sim_{x\rightarrow x^\star}\beta||x-x^\star||^\alpha$ fits this assumption. This is also the case for  more complicated functions such as the one illustrated in Figure~\ref{fig:regretrho}. An example of a function and a partitioning that does not satisfy this assumption is the function $f:x\mapsto 1/\ln{x}$ and a standard partitioning of $[0,1)$ because the function decreases too fast around $x^\star = 0$. As observed by Valko~\cite{valko2013stochastic}, this assumption can be weaken to hold only for values of $f$ that are $\eta$-close to $f(x^\star)$ up to an $\eta$-dependent constant in the simple regret. 

Let us note that the set of assumptions made by prior work (Section~\ref{ss:prior}) can be reformulated using solely Assumption~\ref{ass:lashoo}.
For example, for any $f(x)\sim_{x\rightarrow x^\star}\beta||x-x^\star||^\alpha$, one could consider the semi-metric $\ell(x,y) = \beta||x-y||^\alpha$ for which  the corresponding near-optimality dimension defined by Equation~\ref{eq:near-opt} for a standard partitioning  is $d=0$. Yet we argue that our setting provides a more natural way to describe the complexity of the optimization problem for a given hierarchical partitioning. 

Indeed, existing algorithms, that use a hierarchical partitioning of $\mathcal{X}$, like \HOO{}, do not use the full metric information but instead only use the values $\nu$ and $\rho$, paired up with the partitioning. Hence, the precise value of the metric does not impact the algorithms' decisions, neither their performance. What really matters, is how the hierarchical partitioning of $\mathcal{X}$ fits $f$. Indeed, this fit is what we measure.  To reinforce this argument, notice again that any function can be trivially optimized given a perfectly adapted partitioning, for instance the one that associates $x^\star$ to one child of the root. 

Also, the previous analyses tried to provide performance guarantees based only on the metric and~$f$. 
However, since the metric is assumed to be such that \textit{the cells of the partitioning are well shaped}, the large diversity of possible metrics vanishes. Choosing such metric then comes down to choosing only $\nu$, $\rho$, and a hierarchical decomposition of $\mathcal{X}$. Another way of seeing this is to remark that previous works make an assumption on both the function and the metric, and another on both the metric and the partitioning. We underline that the metric is actually there just to create a link between the function and the partitioning. By discarding the metric, we merge the two assumptions into a single one and convert a topological problem into a combinatorial one, leading to easier analysis. 

To proceed, we  define a \emph{new near-optimality dimension}. For any $\nu>0$ and $\rho\in(0,1)$, the near-optimality dimension $d(\nu,\rho)$ of $f$ with respect to the partitioning $\mathcal{P}$ is defined as follows.
\begin{definition}\label{def:d} Near-optimality dimension of $f$ is 
\[d(\rho) \eqdef \inf \left\{d'\in\mathbb{R}^+:\exists C>0, \hspace{1mm}\forall h\geq 0, \hspace{2mm}\mathcal{N}_h(2\nu\rho^h) \le C\rho^{-d'h}\right\}\!,\]
where $\mathcal{N}_h(\varepsilon)$ is the number of cells $\mathcal{P}_{h,i}$ of depth $h$ such that $\sup_{x\in\mathcal{P}_{h,i}} f(x)\ge f(x^\star)-\varepsilon$. 
\end{definition} 

The hierarchical decomposition of the space $\mathcal{X}$ is the only prior information available to the algorithm. The (new) near-optimality dimension is a measure of how well is this partitioning adapted to~$f$. More precisely, it is a measure of the size of the near-optimal set, i.e.,\@ the cells which are such that $\sup_{x\in\mathcal{P}_{h,i}} f(x)\ge f(x^\star)-\varepsilon$. Intuitively, this corresponds to the set of cells that any algorithm would have to sample in order to discover the optimum.

As an example, any $f$ such that $f(x)\sim_{x \to x^\star}||x-x^\star||^\alpha$, for any $\alpha>0$, has a zero near-optimality dimension with respect to the standard partitioning and an appropriate choice of~$\rho$. 
As discussed by~\citet{valko2013stochastic}, any function such that the upper and lower envelopes of $f$ near its maximum are of the same order has a near-optimality dimension of zero for a standard  partitioning of $[0,1]$. 
An example of a function with $d>0$ for the standard partitioning  is in Figure~\ref{fig:regretrho}. Functions that behave differently in different dimensions have also $d>0$ for the standard partitioning. Nonetheless, for some handcrafted partitioning, it is possible to have $d=0$ even for those troublesome functions. 
 
Under our new assumption and our new definition of near-optimality dimension, one can prove the same regret bound for \HOO{} as~\citet{bubeck2011x} and the same can be done for other related algorithms.

\section{The \SHOO{} algorithm}\label{sec:algo}
\subsection{Description of \SHOO{}}
The \SHOO{} algorithm uses, as a subroutine, an optimizing algorithm that \textit{requires the knowledge} of the function smoothness. We use \HOO{}~\cite{bubeck2011x} 
as the base algorithm, but other algorithms, such as \HCT{}~\cite{azar2014online}, 
could be used as well. \SHOO{}, with pseudocode in Algorithm~\ref{alg:poo}, runs several \HOO{} instances in parallel, hence the 
name \textit{\textbf{p}arallel \textbf{o}ptimistic \textbf{o}ptimization}. The number of base  \HOO{}
instances and other parameters are adapted to the budget of evaluations and are automatically decided on the fly. 

 \begin{wrapfigure}{r}{0.55\textwidth}
 \vspace{-1.5em}
\begin{minipage}{0.55\textwidth}
\begin{algorithm}[H]
\begin{algorithmic}
\STATE {\bf Parameters:} $K$, $\mathcal{P}=\{\mathcal{P}_{h,i}\}$ 
\STATE \quad 
 Optional parameters: $\rho_{\max}, \nu_{\max}$
\STATE {\bf Initialization:}
\STATE \quad $D_{\max} \gets \ln K/\ln\left( 1/\rho_{\max}\right)$ 
\STATE \quad$n \gets 0$ \COMMENT{number of  evaluation performed}
\STATE \quad$N \gets 1$ \COMMENT{number of \HOO{} instances}
\STATE \quad$\mathcal{S} \gets \{(\nu_{\max},\rho_{\max})\}$ \COMMENT{set of \HOO{} instances}
\WHILE{computational budget is available}
	\WHILE{$N \le \tfrac{1}{2}D_{\max}\ln\left(n/(\ln n)\right)$}
		\FOR[start new \HOO{}s]{$i\gets1, \dots, N$}
			\STATE $s \gets \left(\nu_{\max},{\rho_{\max}}^{2N/(2i+1)}\right)$
			\STATE $\mathcal{S} \gets \mathcal{S} \cup \{s\}$
			\STATE Perform $\frac{n}{N}$ function evaluation with \HOO{}($s$)
			\STATE Update the average reward $\widehat{\mu}[s]$ of \HOO{}($s$)
		\ENDFOR
		\STATE $n \gets 2n$ 
		\STATE $N \gets 2N$
	\ENDWHILE{\{ensure there is enough \HOO{}s\}}
	\FOR{$s\in\mathcal{S}$}
		\STATE Perform a function evaluation with \HOO{}($s$)
		\STATE Update the average reward $\widehat{\mu}[s]$ of \HOO{}($s$)
	\ENDFOR
	\STATE $n \gets n+N$ 
\ENDWHILE
\STATE $s^\star \gets \text{argmax}_{s\in\mathcal{S}}~\widehat{\mu}[s]$
\STATE {\bf Output:}
A random point evaluated by \HOO{}($s^\star$)
\end{algorithmic}
\caption{\SHOO{}}
\label{alg:poo}
\end{algorithm}
    \end{minipage}
    \vspace{-1em}
\end{wrapfigure}
Each instance of \HOO{} requires two real numbers $\nu$ and $\rho$. Running \HOO{} parametrized with ($\rho, \nu)$ that are far from the optimal one $(\nu_\star, \rho_\star)$\footnote{the parameters $(\nu,\rho)$ satisfying Assumption 1 for which $d(\nu,\rho)$ is the smallest} would cause \HOO{} to underperform. Surprisingly, our analysis of this \textit{suboptimality gap} reveals that it does not decrease too fast as we stray away from $(\nu_\star, \rho_\star)$. This motivates the following observation. If we \textit{simultaneously} run a slew of \HOO{}s  with different $(\nu,\rho)$s, one of them is going to perform decently well. 

In fact, we show that to achieve good performance, we only require $(\ln n)$ \HOO{} instances, where  $n$ is the current number of function evaluations. Notice, that we do not require to know the total number of rounds in advance which hints that we can hope for a \textit{naturally anytime} algorithm.
%

The strategy of \SHOO{} is quite simple: It consists of running $N$ instances of \HOO{} in parallel, that are all launched with different $(\nu, \rho)$s.  At the end of the whole process,
\SHOO{}  selects the instance~$s^\star$ which performed the best and returns one of the points selected by this instance, chosen uniformly at random. Note that just using a doubling trick in \HOO{} with increasing values of $\rho$ and $\nu$ is not enough to guarantee a good performance. Indeed, it is important to keep track of all \HOO{} instances. Otherwise, the regret rate would suffer way too much from using the value of $\rho$ that is too far from the optimal one. 

For clarity, the pseudo-code of Algorithm~\ref{alg:poo} takes $\rho_{\max}$ and $\nu_{\max}$ as parameters but in Appendix~\ref{ss:proper} we show how to set $\rho_{\max}$ and $\nu_{\max}$ \emph{automatically} as functions of the number of evaluations, i.e., $\rho_{\max}\left(n\right)$, $\nu_{\max}\left(n\right)$.
Furthermore, in Appendix~\ref{ss:sharing}, we explain how to share information between the \HOO{} instances which makes the empirical performance  \emph{light-years better}.

Since \SHOO{} is anytime,  the number of instances $N(n)$ is time-dependent and does not need to be known in advance.
In fact, $N(n)$ is increased alongside the execution of the algorithm. 
More precisely, we want to ensure that 
\[N(n) \ge \tfrac{1}{2}D_{\max}\ln\left( n / \ln n\right), \quad
\text{\ where\ } \quad  D_{\max} \eqdef (\ln K)/\ln\left( 1/\rho_{\max}\right).\]

To keep the set of different $(\nu, \rho)$s well distributed,
the number of  \HOO{}s is not increased one by one but instead is doubled when needed.
Moreover, we also require that \HOO{}s run in parallel, perform the same number of function evaluations. Consequently, when we start running new instances, we first ensure to make these instances on par with already existing ones in terms of number of evaluations. 

Finally, as our analysis reveals, a good choice of parameters $(\rho_i)$ is not a uniform grid on\,$[0,1]$. 
Instead, as suggested by our analysis, we require that $1/\ln(1/\rho_i)$ is a uniform grid on\,$[0,1/\ln(1/\rho_{\max})]$. As a consequence, we add \HOO{} instances in batches 
such that  $\rho_i = {\rho_{\max}}^{N/i}$.

\subsection{Upper bound on \SHOO{}'s simple regret}

\SHOO{} does not require the knowledge of a $(\nu, \rho)$ verifying Assumption~\ref{ass:lashoo} and\footnote{note that several possible values of those parameters are possible for the same function} yet we prove that it achieves a performance close\footnote{up to a logarithmic term $\sqrt{\ln n}$ in the simple regret} to the one obtained by \HOO{} using the best parameters $(\nu_\star,\rho_\star)$. This result solves the open question of~\citet{valko2013stochastic}, whether the stochastic optimization of $f$ with unknown parameters $(\nu, \rho)$  when $d>0$ for the standard partitioning is possible.

 \begin{theorem}\label{thm:poo}
Let~$R_n$ be the simple regret of \SHOO{} at step~$n$. 
For any~$(\nu,\rho)$ verifying Assumption~\ref{ass:lashoo} such that $\nu \le \nu_{\max}$ and $\rho \le \rho_{\max}$ there exists $\kappa$ such that for all $n$
\[\mathbb{E}[R_n] \leq \kappa\cdot\left(\left(\ln^2n\right)/n\right)^{1/(d(\nu,\rho) + 2)}.\]

Moreover, $\kappa = \alpha \cdot D_{\max}{\left(\nu_{\max}/\nu_\star\right)}^{D_{\max}}$, where 
$\alpha$ is a constant independent of $\rho_\text{max}$ and $\nu_\text{max}$. 

\end{theorem}
We  prove Theorem~\ref{thm:poo} in the Appendix~\ref{ss:sketch} and~\ref{sec:appendix}.
Notice that Theorem~\ref{thm:poo}  holds for any $\nu \le \nu_{\max}$ and $\rho \le \rho_{\max}$ and in particular for the parameters $(\nu_\star, \rho_\star)$ for which $d(\nu, \rho)$ is minimal as long as $\nu_\star \le \nu_{\max}$ and $\rho_\star \le \rho_{\max}$. In Appendix~\ref{ss:proper}, we show how to make $\rho_{\max}$ and $\nu_{\max}$ \emph{optional}.

To give some intuition on $D_{\max}$, it is easy to prove that it is the attainable upper bound on the near-optimality dimension of functions verifying Assumption~\ref{ass:lashoo} with $\rho \le \rho_{\max}$.  Moreover, any function of $[0,1]^p$, Lipschitz for the Euclidean metric, has $ (\ln K)/\ln\left( 1/\rho\right) = p$ for a standard partitioning. 

The \SHOO{}'s performance should be compared to the simple regret of \HOO{} run with the best parameters~$\nu_\star$ and~$\rho_\star$, which is of order
\[\cO\left(\left(\left(\ln n\right)/n\right)^{1/(d(\nu_\star,\rho_\star) + 2)}\right)\!.\]
Thus \SHOO{}'s performance is only a factor of $\cO ( \left( \ln n\right)^{1/(d(\nu_\star,\rho_\star) + 2)})$ away from the optimally fitted \HOO{}.
Furthermore, our simple regret bound for \SHOO{} is slightly better than the known simple regret bound for \StoSOO{}~\cite{valko2013stochastic} in the case when $d(\nu,\rho)=0$ for the same partitioning, i.e., $\mathbb{E}[R_n] = \cO\left(\ln n/\sqrt{n}\right).$ With our algorithm and analysis, we generalize this bound for any value of~$d\geq 0$. 

Note that we only give a simple regret bound for \SHOO{} whereas \HOO{} ensures a bound on both the cumulative and simple regret.\footnote{in fact, the bound on the simple regret is a direct consequence of the bound on the cumulative regret~\cite{bubeck2011pure}} Notice that since \SHOO{} runs several \HOO{}s  with non-optimal values of the $(\nu,\rho)$ parameters, this algorithm explores much more than optimally fitted \HOO{}, which dramatically impacts the cumulative regret. As a consequence, our result applies to the simple regret only. 

\section{Experiments}\label{sec:exp}

We ran experiments on the function plotted in Figure~\ref{fig:regretrho} for \HOO{} algorithms with different values of~$\rho$ and the 
\SHOO{}\footnote{code available at \url{https://sequel.lille.inria.fr/Software/POO}} algorithm for $\rho_\text{max} = 0.9$. 
This function, as described in Section~\ref{sec:intro}, has an upper and lower envelope that are not of the same order and therefore has $d>0$ for a standard partitioning.

\begin{figure}
\begin{center}
\vspace{-1em}
\includegraphics[width=0.495\columnwidth]{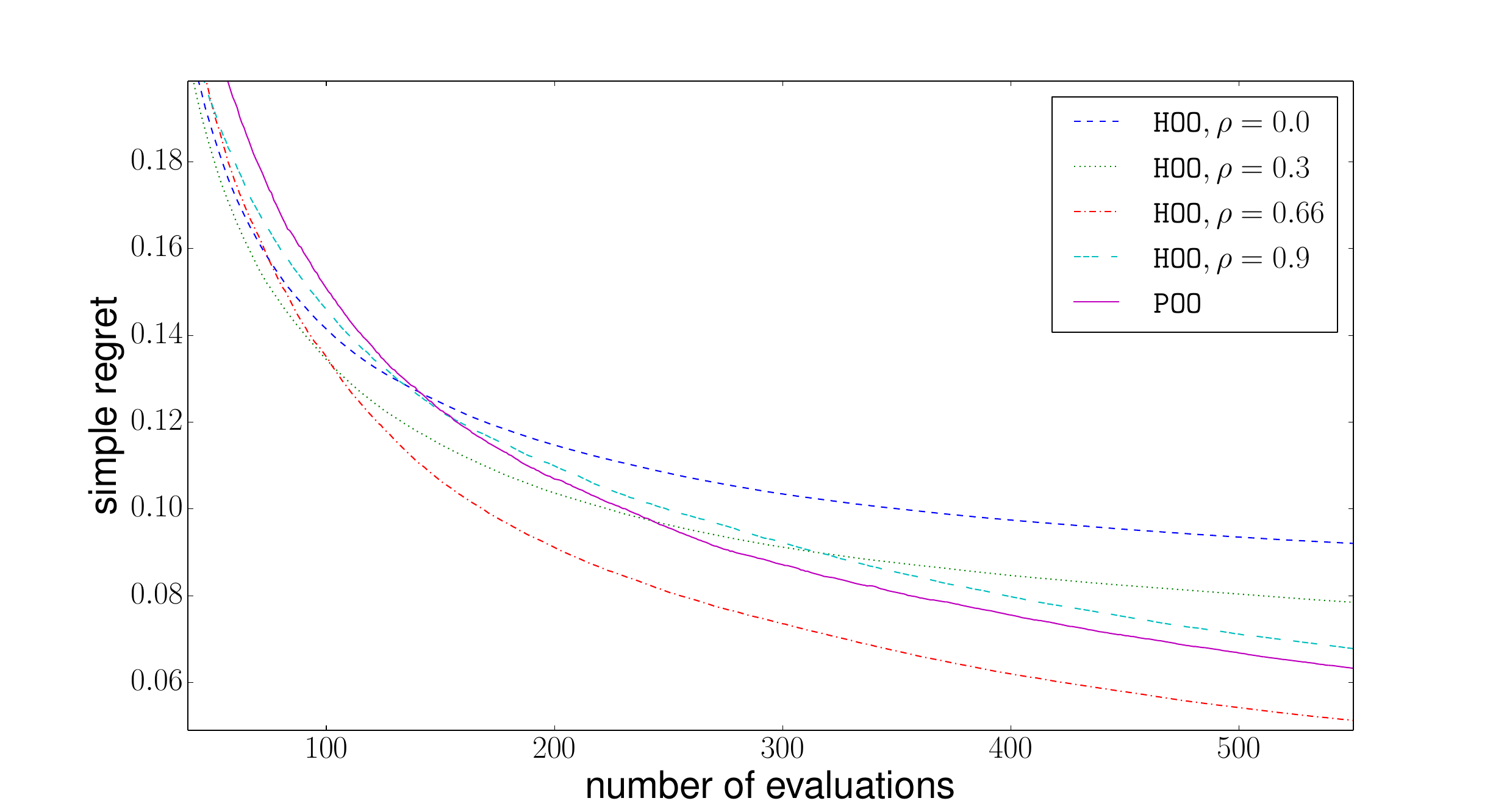}
\includegraphics[width=0.495\columnwidth]{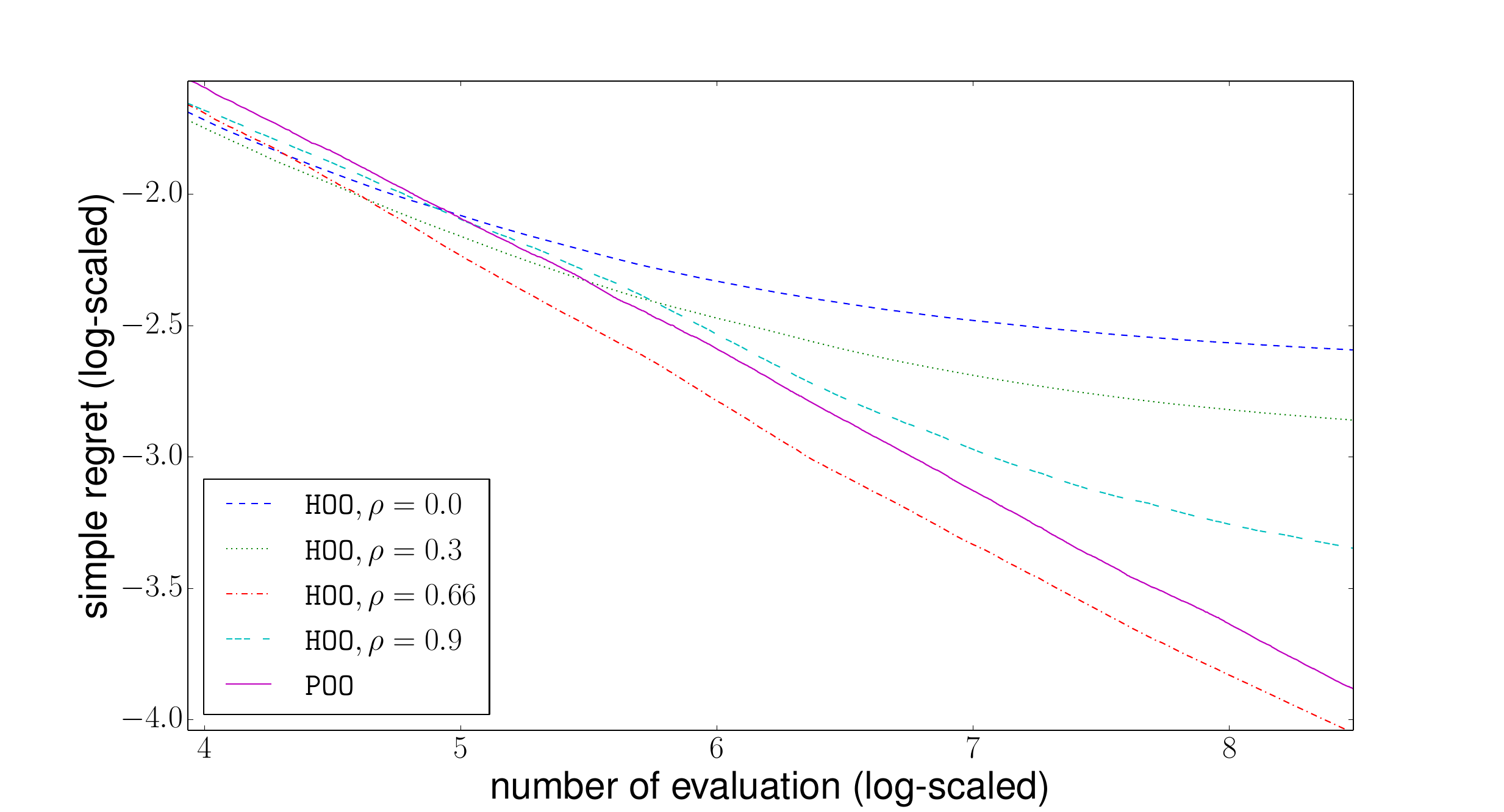}
\vspace{-1em}
\caption{Simple regret of \SHOO{} and \HOO{} run for different values of $\rho$. }
\label{fig:regret}
\vspace{-1em}
\end{center}
\end{figure}

In Figure~\ref{fig:regret}, we show the simple regret of the algorithms as a function of the number of evaluations.  In the figure on the left, we plot the simple regret after  500 evaluations. In the right one, we plot the simple regret after 5000 evaluations in the log-log scale, in order to see the trend better. The \HOO{} algorithms return a random point chosen uniformly among those evaluated. \SHOO{} does the same for  the best empirical instance of  \HOO{}. We compare the algorithms according to the expected simple regret, which is the difference between the optimum and the expected value of function value at the point they return. We compute it as the average of the value of the function for all evaluated points. 
While we did not investigate possibly different heuristics, we believe that returning the deepest evaluated point  would give a better empirical performance. 

As expected, the \HOO{} algorithms using values of $\rho$ that are too low, do not explore enough and become quickly stuck in a local optimum. This is the case for both  \UCT{} (\HOO{} run for $\rho = 0$) and \HOO{} run for $\rho=0.3$. The \HOO{} algorithm using $\rho$ that is too high waste their budget on exploring too much. This way, we empirically confirmed that the performance of the \HOO{} algorithm is greatly impacted by the choice of this $\rho$ parameter for the function we considered. In particular, at $T=500$, the empirical simple regret of \HOO{} with $\rho=0.66$ was a half of the simple regret of  \UCT{}. 

In our experiments, \HOO{} with $\rho = 0.66$ performed the best which is a bit lower than what the theory would suggest, since $\rho_\star = 1/\sqrt{2} \approx 0.7$. The performance of \HOO{} using this parameter is almost matched by \SHOO{}. This is surprising, considering the fact the \SHOO{} was simultaneously running $100$ different \HOO{}s. It shows that carefully sharing information between the instances of \HOO{}, as described and justified in Appendix~\ref{ss:sharing}, has a major impact on empirical performance. Indeed, among the $100$ \HOO{} instances, only two (on average) actually needed a fresh function evaluation, the $98$ could reuse the ones performed by another \HOO{} instance.

\section{Conclusion}\label{sec:con}
We presented \SHOO{}, a parallel optimistic optimization algorithm for black-box optimization of noisy functions with unknown smoothness. \SHOO{} performs almost as well as the best known algorithms that require the knowledge of smoothness, with a simple regret that is at most a factor of $\sqrt{\ln n}$ away. Our analysis applies to a broader class of functions than previously considered, including hard instances with nonzero near-optimality dimension.

\paragraph{Acknowledgements}
\label{sec:Acknowledgements}
The research presented in this paper was supported by French Ministry of
Higher Education and Research, Nord-Pas-de-Calais Regional Council, a doctoral grant
of \'Ecole Normale Sup\'erieure in Paris, Inria and Carnegie Mellon University associated-team project EduBand, and French National Research Agency project ExTra-Learn (n.ANR-14-CE24-0010-01).
\clearpage

\bibliographystyle{plainnat}
\bibliography{perfect_biblio,../../../library}
\clearpage
\appendix

\section{Proof sketch of Theorem~\ref{thm:poo}}
\label{ss:sketch}
In this part we give the roadmap of the proof. The full proof is in Appendix~\ref{sec:appendix}. 
\paragraph{First step}
For any choice of $\rho_\star$ verifying Assumption~\ref{ass:lashoo} and any suboptimal $\rho$ such that 
$$0 < \rho_\star \le \rho < 1,$$ we bound the difference of near-optimality dimension, 
\[
d\left(\rho\right) - d\left(\rho_\star\right) \le \ln K\left(\frac{1}{\ln\left(1/\rho\right)} - \frac{1}{\ln\left(1/\rho_\star\right)}\right)\!
\]
and deduce that 
\[\min_{i:\rho_i \ge \rho_{\star}} \left[d(\rho_i) - d(\rho_\star) \right] \le \frac{D_{\max}}{N}\cdot\]
\paragraph{Second step}
By simultaneously running a large number of \HOO{} instances, we ensure that for all $\rho_\star\le\rho_{\max}$, one of them uses a $\rho$ close to  $\rho_\star$ and therefore suffers a low regret. On the other hand, simultaneously running a large number of \HOO{}s has a cost, as more evaluations need to be done at each step, one for each \HOO{}. We optimize this tradeoff to deduce the following good choice of $\delta$, which is the maximum distance $\left|d\left(\rho_i\right) - d(\rho_j)\right|$, where $i$ and $j$ are two consecutive \HOO{}s. 
\[\delta = \cO\left(\ln\left(t/\ln t\right)\right).\]

\paragraph{Third step}
Using the result of the second step, we can compute the simple regret $R^{\rho}_n$  of the \HOO{} instance running with the parameter $\bar\rho>\rho_\star$, which is the closest to $\rho_\star$. Note that, as \SHOO{} is running, the instance it chooses may change over time and so $\bar\rho$ depends on $n$. 

We prove that there exists a constant $\alpha>0$ such that for all $n$, $\nu_{\max}>0$, and $\rho_{\max}<1$,
\[R^{\rho}_n  \le \alpha \cdot D_{\max}{\left(\nu_{\max}/\nu_\star\right)}^{D_{\max}}\left(\left(\ln^2 n\right)/n\right)^{1/(d(\bar\rho)+2)}.\] 
\paragraph{Fourth step}
At the end of the algorithm, we empirically determine which \HOO{} performed the best. However, this best empirical instance may not be the instance running with  $\rho$ closest to the optimal unknown $\rho_\star$. 
Nonetheless, we prove that this error is small enough such that it only impacts the simple regret by a constant factor.

\section{Full proof of Theorem~\ref{thm:poo}}
\label{sec:appendix}


\subsection{First step}

%
%
%
%
%
We show that for any choice of $\rho_\star$ verifying Assumption~\ref{ass:lashoo} and any  $\rho$ such that $0<\rho_\star\le\rho < 1$,
\begin{align*}
d\left(\rho\right) - d\left(\rho_\star\right) \le \ln K\left(\frac{1}{\ln\left(1/\rho\right)} - \frac{1}{\ln\left(1/\rho_\star\right)}\right)\!\cdot
\end{align*}
We start by defining $\mathcal{I}_{h}(\varepsilon)$ as the set of cells of depth $h$ which are $\varepsilon$-near-optimal, 
\begin{align*}
\mathcal{I}_{h}\left(\varepsilon\right) \eqdef \left\{ i : \sup_{x\in\mathcal{P}_{h,i}} f(x) \ge f(x_\star) - \varepsilon \right\}\!\cdot
\end{align*}
$\mathcal{N}_h(\epsilon)$, defined in Section~\ref{sec:intro}, is then equal to the cardinality of $\mathcal{I}_{h}(\epsilon)$. Notice that if a cell $(h,i)$ is $\varepsilon$-near-optimal then all of its antecedents are also $\varepsilon$-near-optimal. Therefore, for any $\varepsilon$ and $h' > h$, the cells in $\mathcal{I}_{h'}(\varepsilon)$ are descendants of the cells in $\mathcal{I}_{h}(\varepsilon)$. 

Since the number of descendants at depth $h'$ of a cell at depth $h' > h$ is bounded by $K^{h'-h}$ we bound the cardinality $\mathcal{N}_h(\epsilon)$ of $\mathcal{I}_{h'}(\varepsilon)$,
\begin{align*}
\forall \varepsilon,\forall h'>h, \quad \mathcal{N}_ {h'}(\varepsilon) \le K^{h'-h}
\mathcal{N}_{h}(\varepsilon). 
\end{align*}
By definition of the near-optimality dimension, we know that for any $\nu>0$ and $\rho_\star\in(0,1)$,  there exists $C$ such that for all~$h$,
\begin{align*}
\mathcal{N}_{h}\left(2\nu\rho^{h}\right) \le C\rho^{-d\left(\rho\right)h}.
\end{align*}
We define $C(\nu, \rho)$ as the smallest $C$ verifying the above condition. 

For any $0 < \nu_\star < \nu$, $0<\rho_\star<\rho<1$ and any integer $h\ge h_{\min} \eqdef \ln(\nu/\nu_\star)/\ln(1/\rho)$ let us define $h_\star$ as the greatest integer such that $\nu\rho^{h} < \nu_\star\rho_\star^{h_\star}$. From this definition, we get $\nu\rho^{h} \ge \nu_\star\rho_\star^{h_\star + 1}$ from which we deduce that
\[h_\star \ge h\cdot\frac{\ln\rho}{\ln\rho_\star}+\frac{\ln\nu-\ln\nu_\star}{\ln\rho_\star}-1,\]   and then
\begin{align*}
 h-h_\star \le h_\star\ln\rho_\star\left(\frac{1}{\ln\rho}-\frac{1}{\ln\rho_\star}\right)+\frac{\ln\rho_\star+\ln\nu_\star-\ln\nu}{\ln\rho}\cdot
\end{align*}

Since $\mathcal{N}_{h}(\varepsilon)$ is not increasing in $\varepsilon$,  $\nu\rho^{h} < \nu_\star\rho_\star^{h_\star}$ implies 
\[\mathcal{N}_{h}(2\nu\rho^{h}) \le \mathcal{N}_{h}(2\nu_\star\rho_\star^{h_\star}).\]
We now put everything together to obtain
\begin{align*}
\mathcal{N}_{h}(2\nu\rho^{h}) &\le \mathcal{N}_{h}(2\nu_\star\rho_\star^{h_\star})\\
&\le K^{h-h_\star} \mathcal{N}_{h_\star}(2\nu_\star\rho_\star^{h_\star})\\
&\le K^{(\ln\rho_\star+\ln\nu_\star-\ln\nu)/ \ln\rho+h_\star\ln\rho_\star\left(1/\ln\rho-1/\ln\rho_\star\right)}C(\nu_\star,\rho_\star)\rho_\star^{-d(\rho_\star)h_\star}\\
&\le  C(\nu_\star,\rho_\star) K^{(\ln\rho_\star+\ln\nu_\star-\ln\nu)/ \ln\rho} \rho_\star^{-h_\star\left[d(\rho_\star) + \ln K\left(1/\ln\left(1/\rho\right) - 1/\ln\left(1/\rho_\star\right)\right)\right]}.
\end{align*}

From $\nu\rho^{h} < \nu_\star\rho_\star^{h_\star}$ and $\nu_\star < \nu$ we get $\rho^{-h} > \rho_\star^{-h_\star}$ and therefore
\begin{align*}
\mathcal{N}_ {h}(2\nu\rho^{h}) \le C(\nu_\star,\rho_\star) K^{(\ln\rho_\star+\ln\nu_\star-\ln\nu)/ \ln\rho}\rho^{-h\left[d(\rho_\star) + \ln K\left(1/\ln\left(1/\rho\right) - 1/\ln\left(1/\rho_\star\right)\right)\right]}.
\end{align*}
We just proved that there exists $C$ such that for all $h>0$
\begin{align*}
\mathcal{N}_ {h}(2\nu\rho^{h}) \le C \rho^{-h\left[d(\rho_\star) + \ln K\left(1/\ln\left(1/\rho\right) - 1/\ln\left(1/\rho_\star\right)\right)\right]}.
\end{align*}
By taking 
\begin{align*}
C \eqdef \max \left(C(\nu_\star,\rho_\star) K^{(\ln\rho_\star+\ln\nu_\star-\ln\nu)/ \ln\rho} , K^{h_{\min}}\right)\!,
\end{align*}
we deduce by the definition of the near-optimality dimension the following bound
\begin{align*}
d(\rho) \le d(\rho_\star) + \ln K\left(\frac{1}{\ln\left(1/\rho\right)} -\frac{1}{\ln\left(1/\rho_\star\right)}\right)\!\cdot
\end{align*}
We can now deduce that \SHOO{} should use $\rho_i$ parameters that satisfy
\[\frac{1}{\ln\left(1/\rho_i \right)} \eqdef \frac{i}{N}\frac{1}{\ln\left( 1/ \rho_{\max}\right)}\CommaBin\]
where $N$ is the total number of \HOO{} instances run and $i\in\{1,\dots,N\}$. 

We now define $\bar\rho$ as the closest $\rho_i$ to $\rho_\star$ used by an existing \HOO{} instance, such that $\rho_i>\rho_\star$. 
\[\bar\rho \eqdef \argmin_{\rho_i \ge \rho_{\star}} \left[d\left(\rho_i\right) - d\left(\rho_\star\right) \right].\]
Since we assumed that $\rho_\star < \rho_{\max}$, we know that
\[d(\bar\rho)-d(\rho_\star)\le\frac{D_{\max}}{N}, 
\text{\quad with \quad }
D_{\max} \eqdef  (\ln K)/\ln\left( 1/\rho_{\max} \right).\]

\subsection{Second step}

Let us now compute the optimal number of $N$ instances to run in parallel. We bound the logarithm of the simple regret $R^{\nu, \rho}_t$ of a single \HOO{} instance using parameters $\nu$ and $\rho$ after this particular instance performed $t$ function evaluations. In particular, we bound the simple regret by a linear approximation for $\rho\sim\rho_\star$.  In the following, $\beta$ is a numerical constant coming from the analysis of \HOO{}~\cite{bubeck2011x}. For all $t>0$, we have\begin{align*}
\ln R^{\nu,\rho}_t & \le \ln\beta + \frac{\ln C(\nu,\rho)}{2+d(\rho)} - \frac{\ln\left(t/\ln t \right)}{2+d(\rho)} \\ 
& =  \ln\beta + \frac{\ln C(\nu,\rho)}{2+d(\rho)} -  \frac{\ln\left(t/\ln t\right)}{2+d(\rho_\star)} \cdot \frac{1}{1+\left(d\left(\rho\right)-d\left(\rho_\star\right)\right)/\left(2+d\left(\rho_\star\right)\right)} \\
& \le \ln\beta + \frac{\ln C(\nu,\rho)}{2+d(\rho)} -\frac{\ln\left(t/\ln t\right)}{2+d(\rho_\star)} \cdot \left(1-\frac{d(\rho)-d(\rho_\star)}{2+d(\rho_\star)}\right)\!\cdot  
\end{align*}

After $n$ function evaluations by \SHOO{}, each instance performed at least $t = \lfloor n/N \rfloor$ function evaluations. We can now bound the simple regret $R^{\SHOO{},\nu,\bar\rho}_n$ of the \HOO{} instance using $\nu$ and $\bar\rho$ after $n$ evaluations performed by all the instances
\begin{align}\label{eq:regretPOOHOO}
\ln R^{\SHOO{},\nu, \bar\rho}_n  \leq \ln\beta +  \frac{\ln C(\nu,\bar\rho)}{2+d(\bar\rho)} +  \ln\left(\frac{\ln \lfloor n/N \rfloor}{\lfloor n/N \rfloor}\right)\left(\frac{1}{2+d(\rho_\star)} - \frac{D_{\max}/N}{(2+d(\rho_\star))^2}\right)\!\cdot
\end{align}
Optimizing this upper bound for $N$ leads to the following choice of $N$,
\[N\sim \tfrac12 D_{\max}\ln\left(n/ \ln n\right).\]

Therefore, in \SHOO{} we choose to ensure $N\ge \tfrac12 D_{\max}\ln\left( n/ \ln n\right)$. 

If the time horizon was known in advance, $N$ could be any integer. Nevertheless, since the algorithm is anytime, all the previous \HOO{} instances have to be kept and new instances need to be added in between. Therefore, we restrict $N$ to be of the form 
$2^i$, for $i\in\mathbb{N}$. 

As a consequence of this choice, $N$ can be at most 2 times its lower bound and therefore 
\[\tfrac12 D_{\max}\ln\left(n/ \ln n\right) \le N \le D_{\max}\ln\left(n/\ln n\right).\]

\subsection{Third step}
Using our choice of $N$, we can  bound the simple regret of the \HOO{}  instance using $\bar\rho$. We proceed by separately bounding each of the terms in Equation~\ref{eq:regretPOOHOO}.
\begin{align*}
\frac{\ln C(\nu,\bar\rho)}{2+d(\bar\rho)} &\le \frac{1}{2+d(\rho_\star)}\ln C(\nu,\bar\rho) \\
& \le \frac{1}{2+d(\rho_\star)}\ln\max\left(
C(\nu_\star, \rho_\star)K^{\left(\ln\rho_\star+\ln\nu_\star-\ln\nu\right)/\ln\bar\rho}, K^{h_{\min}}\right) \\
&\le \frac{1}{2+d(\rho_\star)}\! \max  \! \left( \ln C(\nu_\star\rho_\star) \!+\!\ln K\left(\frac{\ln1/\rho_\star}{\ln1/\bar\rho}+\frac{\ln\left(\nu/\nu_\star\right)}{\ln1/\rho}\right), \ln \left[K^{\ln(\nu/\nu_\star)/\ln(1/\rho)}\right]\right)\\
&\le \frac{1}{2+d(\rho_\star)}\! \max \!\left( \ln C(\nu_\star\rho_\star) \!+ \!
\max\left(\!\frac{\ln K \ln\rho_\star D_{\max}}{N},2\!\right)
\!+\!\frac{\ln K\ln\frac{\nu_{\max}}{\nu_\star}}{\ln1/\rho}, {D_{\max}}\ln \frac{\nu}{\nu_\star} \right)\\
&\le \gamma  + \frac{D_{\max}}{2+d(\rho_\star)}\ln \left(\nu_{\max}/\nu_\star\right)
\end{align*}

In the last expression, $\gamma$ is a quantity independent of $\nu_{\max}$, $\rho_{\max}$, and $N$. 

We now use $N \le D_{\max}\ln\left(n/\ln n\right)$ to get

\[\ln\left(\frac{\ln \lfloor n/N \rfloor}{\lfloor n/N \rfloor}\right) \le \ln\left(D_{\max}\ln n \ln\left(n/\ln n\right)/n\right).\]

To bound the last term, we use $\tfrac12 D_{\max}\ln\left(n/ \ln n\right) \le N$ to get
\begin{align*}
-\ln\left(\frac{\ln \lfloor n/N \rfloor}{\lfloor n/N \rfloor}\right)\frac{D_{\max}/N}{(2+d(\rho_\star))^2} &\le  \ln\left(
\frac{1}{D_{\max}}\cdot\frac{n}{\ln n}\cdot\frac{1}{\ln\left(n/\ln n\right)}
\right) \frac{1}{2\ln\left(n/\ln n\right)}  \le 2.
\end{align*}
We can finally bound the simple regret $R^{\SHOO{},\bar\rho}_n$ of the \HOO{} instance using $\bar\rho$ after $n$ function evaluations overall. 
Combining the results above, we know that for all $n$, $\nu_{\max},$ and $\rho_{\max}$,
\begin{align*}
R^{\SHOO{},\bar\rho}_n \le \beta\exp(\gamma + 2)\left(D_{\max}\left(\nu_{\max}/\nu_\star\right)^{D_{\max}}\left(\ln n\right)
\ln\left(n/\ln n\right)/ n\right)^{1/(2+d(\bar\rho))}.
\end{align*}

We bound $\ln\left(n/\ln n\right)$ by $\ln n$ to get the following bound. There exists $\alpha$ that is independent of $\rho_\text{max}$ and $\nu_\text{max},$ such that
\[R^{\SHOO{},\bar\rho}_n  \le \alpha \cdot D_{\max}{\left(\nu_{\max}/\nu_\star\right)}^{D_{\max}}\left(\left(\ln^2 n\right)/n\right)^{1/(d(\bar\rho)+2)}.\] 

\subsection{Fourth step}
\label{ss:proof4}


Let $(X_{i,j})_{i\le n, j\le N}$ be a family of points in $\mathcal{X}$ evaluated by \SHOO{}. We denote $\widehat{f}(X_{i,j})$ the noisy evaluation at $X_{i,j}$ and $f(X_{i,j}) = \mathbb{E}[\widehat{f}(X_{i,j})]$. We also define:
\begin{alignat*}
{\mu}\mu_j &\eqdef \frac{1}{n}\sum_{i=1}^n f(X_{i,j}) 
\qquad &\widehat{\mu}_j &\eqdef  \frac{1}{n}\sum_{i=1}^n \widehat{f}(X_{i,j})  \\
\widetilde\jmath & \eqdef \argmax_{1\le j \le N} \mu_j 
\qquad & 
\widehat{\jmath} &\eqdef \argmax_{1\le j \le N} \widehat{\mu}_j 
\end{alignat*}

By Hoeffding-Azuma inequality for martingale differences, for any $\Delta > 0,$
\[ \mathbb{P}\left[\bigg|  \sum_{i=1}^n \widehat{f}(X_{i,j}) - f(X_{i,j}) > n\Delta\bigg|\right] \le 2\exp\left(-\cfrac{2(n\Delta)^2}{n}\right)\!\cdot\]
Therefore 
\[ \mathbb{P}\left[|\widehat{\mu}_j - \mu_j > \Delta|\right] \le 2\exp\left(-2n\Delta^2\right)\!.\]
As we have 
\[\forall x\ge0, x\cdot\exp\left(-2nx^2\right) \le \frac{e^{-2}}{2\sqrt{n}}\CommaBin\]
we can now integrate $\exp\left(-2n\Delta^2\right)$ over $\Delta\in[0,1]$ to get
\[ \mathbb{E}[|\widehat{\mu}_j - \mu_j|] \le\cfrac{e^{-2}}{\sqrt{n}}\cdot\]
Now consider 
\[\EE{\mu_{\widetilde\jmath} - \mu_{\widehat{\jmath}}} = \EE{\mu_{\widetilde\jmath} - \widehat{\mu}_{\widetilde{\jmath}}} + 
\EE{\widehat{\mu}_{\widetilde\jmath} - \widehat{\mu}_{\widehat{\jmath}}} + 
\EE{\widehat{\mu}_{\widehat{\jmath}} - \mu_{\widehat{\jmath}}}\!.\]
Notice that the first and last term are both bounded by $e^{-2}/\sqrt{n}$ 
and the middle term is negative.  Furthermore notice, that we used the 
cumulative regret guarantee of \HOO, the recommendation strategy of 
each instance, and also of \SHOO.
Finally, taking a union bound over the $N$ variables $\mu_j$ we get 
\[ \EE{\mu_{j_\star} - \mu_{\widehat{\jmath}}}  \le \frac{e^{-2}N}{\sqrt{n}}\cdot\]

As $N = o\left(\ln n\right)$, we conclude that this additional term is negligible with respect to 
\[
\left(\ln n \ln\left(n/\ln n\right)/n \right)^{1/(2+d(\rho_\star))}.
\]

\section{Increasing sequence for $\rho_{\max}$ and $\nu_{\max}$}
\label{ss:proper}
Besides the number $K$ of children for each cell, \SHOO{} needs two parameters, $\rho_{{\max}}\in(0,1)$ and $\nu_{{\max}}>0$. Theorem~\ref{thm:poo}  states that \SHOO{} run with those parameters performs almost as well as the best instance of \HOO{} run with $\nu\leq \nu_{\max}$ and $\rho\leq \rho_{\max}$, i.e., corresponding to the near-optimality dimension  $\min\{ d(\nu,\rho), \nu\leq \nu_{\max}, \rho\leq \rho_{\max}\}$.

Therefore, the larger the values $\rho_{\max}$ and $\nu_{\max}$ used by \SHOO{}, the wider the set of \HOO{} instances that we can compete with.
Nevertheless, large values of $\rho_{\max}$ and  $\nu_{\max}$ impact the performance by a multiplicative constant of order $D_{\max}{\nu_{\max}}^{D_{\max}}$. This tradeoff between performance and size of our comparison class is unfortunate but unavoidable.


In practice, as we strive for an algorithm that does not require the knowledge of the smoothness we may increase the values of $\rho_{\max}(n)$ and $\nu_{\max}(n)$ with the number of evaluations $n$, so that the class of functions covered by \SHOO{} gets bigger with the numerical budget. Nevertheless, the increase should be slow enough so that we do not compromise the performance. In particular, we will require that $\nu_{\max}(n)^{D_{\max}(n)}$ does not increase too fast. 
In fact, any sequence $\rho_{\max}(n)$ converging to 1 and $\nu_{\max}(n)$ diverging to infinity impacts the simple regret by an additive term which is the smallest time $n$ such that $\rho^\star < \rho_{\max}(n)$ and $\nu^\star < \nu_{\max}(n)$, i.e., the first time the assumptions are verified.
A slowly increasing sequence means a smaller impact on the simple regret rate but a higher additive term (a constant independent of $n$).  Any sensible choice of increasing sequence $\rho_{\max}(n)$ and $\nu_{\max}(n)$,  impacting the rate by only a subpolynomial factor, is a valid choice. 




Algorithm~\ref{alg:poo} is described using constant $\rho_{\max}$ and $\nu_{\max}$ for clarity, but its implementation is easily modifiable to deal with increasing values of these two parameters while preserving the anytime property of the algorithm, as follows. At any time, all the \HOO{} instances must use the same $\nu_{\max}$ parameter. On the other hand, considering $\rho_{\max}$, the value of $D_{\max}$ has to be increased such that the already running \HOO{} instances stay relevant. One way to do that is to increase $D_{\max}$ as $D_{\max}(N+1)/N $ and run an additional \HOO{} instance. An alternative solution is to perform, each time when needed, the following increment $\rho_{\max} \gets \sqrt{\rho_{\max}}$ and run $N$ additional \HOO{} instances with parameters ${\rho_{\max}}^{2N/i}$, for $i\in\{1,\dots,N\}$. 
 \section{Information sharing among parallel runs}
\label{ss:sharing}
Since we run several instances of \HOO{} on the same partitioning of $\mathcal{X}$, we may think of \emph{sharing the samples} 
among them, in order to decrease the estimation error.  However, this needs to be done carefully in order to avoid adding unwanted bias in the estimation of the $U$ values in the \HOO{} instances.  
Ideally, each \HOO{} instance would reuse all function evaluations acquired by all other instances. Unfortunately, this solution would not easily come with theoretical guarantees, as this would reduce artificially the confidence intervals at some cells and introduce \emph{search bias}. 

 Instead, whenever a \HOO{} instance requires a function evaluation, we perform a \emph{look-up} to find out whether another \HOO{} instance has already evaluated $f$ at this point. In affirmative, then instead of evaluating the function at this point again, we simply \emph{reuse} the sample. This way, \HOO{} instances are \emph{not given access to samples they never asked for}.  
 However, the empirical simple regrets of \HOO{}s becomes correlated with each other. This is not a problem because in \ref{ss:proof4}, we do not assume the independence between empirical means of \HOO{}s, only the independence of rewards within each instance---which still holds. 
Therefore with this modification, our theoretical guarantees continue to apply. Note that if all the instances share all their rewards, then they are all equivalent and there is no mistake possible. Then one can show, that the worst case is when no rewards are shared and  the error due to choosing the wrong instance actually decreases when the information is shared. 

Finally, we want to stress that sharing information is extremely important in practice, as our experiments reveal. Since the number of \HOO{} instances can be very large\footnote{even though it scales only as $\ln n$ with the number of evaluations $n$, it does not scale well with $\rho_{\max}$} one could expect the performance of \SHOO{} to be pitiful. However, as the vast majority of the function evaluations are in practice shared, \SHOO{} performs almost as well as \HOO{} fitted with the best parameters.
Summing up, although the performance bound on the simple regret with this modification is the same, empirical performance 
\emph{improves tremendously}.

\section{Zero-quality functions}
\label{ss:quality}

For any $\rho \in (0,1)$, we construct a  locally Lipschitz function with a rate $\rho$ and a constant $\nu = 1$ that \SHOO{} can provably optimize and its \emph{quality}, as defined by Definition~\ref{def:quality}, is zero. In order to properly define the quality, 
we use the uniform distribution on $[0,1]$ to sample from a node of the partitioning. 

\begin{definition}[\citet{slivkins2011multi-armed}]
\label{def:quality}
The quality is the largest $q\in(0,1)$ such that for each subtree $v$ containing the optimum, there
exist nodes $u$ and $u'$ such that  $\mathbb{P}(u|v)$ and  $\mathbb{P}(u'|v)$ are at least $q$ 
and \[|f(u)-f(u')| \geq \tfrac12 \sup_{x,y\in v} \left|f(x)-f(y)\right|\!.\]
\end{definition}

We construct such function $f$ on the interval $[0,1]$, its maximum being attained in $x^\star=0$ with $ f(0) = 0$. 
For any $x\ne0$ we define $f$ as follows. For any $h\ge0$ we define $f$ on $\left(\frac{1}{2^{h+1}}, \frac{1}{2^{h}}\right]$ as

\[ \forall x\in\left(\frac{1}{2^{h+1}}\CommaBin \frac{1+1/(h+1)}{2^{h+1}}\right]\!\CommaBin \quad f(x) = -\rho^h,\]
\[ \forall x\in\left(\frac{1+1/(h+1)}{2^{h+1}}\CommaBin \frac{1}{2^{h}}\right]\!\CommaBin \hspace{2.5em}  f(x) = -\frac{\rho^h}{3}\cdot\]

We also consider the standard partitioning on $[0,1]$. 

The optimal node of depth $h$ corresponds to the interval $\left[0,2^{-h}\right]$. By our definition of $f$, 

\[ \forall x\in\left[0,2^{-h}\right], \hspace{3mm} f(0) - f(x) \le \rho^h\]
\[ f(0) - f\left(\frac{1+1/(h+1)}{2^{h+1}}\right) = \rho^h,\]

from which we conclude that $f$ is locally Lipschitz with rate $\rho$ and therefore can be optimized 
by \SHOO{} with provable finite-time guarantees (Theorem~\ref{thm:poo}).

Now we prove that the quality of this function is zero. Using Definition~\ref{def:quality}, we can do it by showing that there exists no such $q\in(0,1)$, for which there could be a node $v$ along the optimal path with $u$ and~$u'$ verifying $\mathbb{P}(u|v) \ge q$ (and same for $u'$) such that 
\begin{align}
\label{eq:quality}
\sup_{x\in u} f(x) - \sup_{x\in u'} f(x) \ge \sup_{x\in v}\frac{f(0) - f(x)}{2}\cdot
\end{align}
Let $q$ be a real number from $(0,1)$ and consider any $h > 1/q$. We pick $v = \left[0,2^{-h}\right]$. 
\begin{align*} 
 \PP\left(\left\{x\in v: f(x) \le -\frac{\rho^h}{2}\right\}\bigg|v\right) = \\ 
& \hspace{-14em} =  \frac{1}{2}\PP\left(\left\{x\in\left[0,\frac{1}{2^{h+1}}\right]: f(x) \le -\frac{\rho^h}{2}\right\}\bigg|v\right) 
+ \frac{1}{2}\PP\left(\left\{x\in\left[\frac{1}{2^{h+1}}, \frac{1}{2^h}\right]: f(x) \le -\frac{\rho^h}{2}\right\}\bigg|v\right) \\
&\hspace{-14em}=   \frac{1}{2}\PP\left(\left\{x\in\left[0,\frac{1}{2^{h+1}}\right]: f(x) \le -\frac{\rho^h}{2}\right\}\bigg|v\right)  + \frac{1}{2(h+1)}  \\
&\hspace{-14em}\le  \frac{1}{2}\sum_{k=1}^{\infty} \frac{1}{2^k(h+k+1)} + \frac{1}{2(h+1)} \\
&\hspace{-14em}\le  \frac{1}{h+1}< q
\end{align*}
Notice that if $u'$ verifies \eqref{eq:quality}, then $u'$ is included in $\left\{x\in v:  f(x) \le -\rho^h/2\right\}$. Combined with the equation above, we have that 
\[\mathbb{P}(u|v) \le \PP\left(\left\{x\in v: f(x) \le - \rho^h/2\right\}\big|v\right) < q, \] which is a contradiction.  
Since this holds for any $q>0$, we deduce that the quality of $f$ is zero. Yet~$f$ is Lipschitz with rate $\rho\in(0,1)$ and therefore $f$ can be optimized by \SHOO{}.

\end{document}